\documentclass{article} 
\usepackage[final]{colm2025_conference}

\usepackage{microtype}
\usepackage{hyperref}
\usepackage{url}
\usepackage{booktabs}
\usepackage{amsmath}  
\usepackage{amssymb} 
\usepackage{subcaption}
\PassOptionsToPackage{final}{graphicx}
\usepackage{graphicx}

\usepackage{lineno}

\definecolor{darkblue}{rgb}{0, 0, 0.5}
\hypersetup{colorlinks=true, citecolor=darkblue, linkcolor=darkblue, urlcolor=darkblue}

\title{Evaluating Contrast Localizer for Identifying Causal Units\\in Social \& Mathematical Tasks in Language Models}

\author{%
    Yassine Jamaa$^{1,2}$ \quad
    Badr AlKhamissi$^{1}$ \quad
    Satrajit Ghosh$^{*,2}$ \quad
    Martin Schrimpf$^{*,1}$ \\
    $^1$EPFL \quad $^2$MIT \quad $^*$Equal Supervision
}

\begin{document}


\ifcolmsubmission
\fi

\maketitle

\begin{abstract}
This work adapts a neuroscientific contrast localizer to pinpoint causally relevant units for Theory of Mind (ToM) and mathematical reasoning tasks in large language models (LLMs) and vision-language models (VLMs). Across 11 LLMs and 5 VLMs ranging in size from 3B to 90B parameters, we localize top-activated units using contrastive stimulus sets and assess their causal role via targeted ablations. We compare the effect of lesioning functionally selected units against low-activation and randomly selected units on downstream accuracy across established ToM and mathematical benchmarks. Low-activation units sometimes produced larger performance drops than the highly activated ones, and units derived from the mathematical localizer often impaired ToM performance more than those from the ToM localizer. These findings call into question the causal relevance of contrast-based localizers and highlight the need for broader stimulus sets and more accurately capture task-specific units.
\end{abstract}

\section{Introduction}

Recent breakthroughs in LLMs have shown their ability to perform more than language processing tasks, showing their ability to accomplish mathematical, problem-solving tasks \citep{sun_survey_2024, giadikiaroglou_puzzle_2024} and even mimicking social understanding \citep{street_llms_2024}. Although their internal workings are still poorly understood and can be seen as a black box, a growing body of work, such as mechanistic interpretability, has been dedicated to understanding the components that encode specific types of knowledge \citep{geiger_causal_2021, wang_interpretability_2022}. Recent advances in neuroscience—particularly in mapping cognitive networks—have opened new avenues for exploring the internal mechanisms of language models \citep{schrimpf_brain-score_2018, schrimpf_neural_2021}. Unlike the human brain, artificial neural networks offer the unique advantage of allowing targeted perturbations \citep{schrimpf_topographic_2024} to individual units, enabling precise investigations of their functional roles. Central to this emerging direction is the study by \citet{alkhamissi_llm_2024}, which leverages a neuroscientific approach for identifying task-relevant units in LLMs. Drawing on methods traditionally used to localize functional regions in the brain, the authors applied a ``localizer'' paradigm to LLMs. For instance, a language network localizer—typically used in fMRI to identify language-selective areas in the human cortex—successfully pinpointed a subset of model units whose ablation resulted in significant impairments on linguistic tasks. The study further presented preliminary findings for other tasks related to mathematical reasoning and ToM by using localizers used to identify the multiple demand (MD) and ToM brain regions respectively. 

Building on this foundation, we conduct a comprehensive set of experiments using 11 LLMs and 5 VLMs ranging from 3B to 90B parameters to systematically identify and evaluate causal units associated with ToM and MD tasks. To assess the functional significance of these units, we compare the performance impact of ablating them against two baselines: randomly selected units and those with minimal activation during the task. Recognizing the multimodal nature of many real-world problems, we further extend our analysis to include vision-language models (VLMs), exploring whether similar functional localization emerges across modalities.

\section{Related Work}

\textbf{Theory of mind (ToM)} is a specialized brain system — primarily involving regions in the bilateral temporo-parietal junction and along the cortical midline \citep{gallagher_reading_2000, saxe_its_2006} — that becomes active when individuals think about others mental states and understand that they may differ from their own's. It plays a significant role in moral judgment \citep{leslie_acting_2006, sosa_moral_nodate}, anticipation of other's action \citep{baker_action_2009, baker_rational_2017} and in understanding sarcasm \citep{spotorno_neural_2012, hsu_two_2013, bischetti_pragmatics_2023}.

\textbf{Multiple Demand (MD)} is a brain system closely linked to working memory, cognitive control, and attention, all of which are critical for goal-directed behavior \citep{assem_domain-general_2020, woolgar_fluid_2010} and includes bilateral frontal and parietal regions, medial prefrontal areas, and posterolateral inferior temporal regions \citep{cole_cognitive_2007}. These areas become active with stronger activation observed in more difficult tasks \citep{duncan_common_2000, fedorenko_broad_2013, shashidhara_individual-subject_2020}, a pattern that extends to various task types, including learning new tasks, memory, math problems, and logic puzzles \citep{duncan_common_2000, fedorenko_broad_2013, shashidhara_individual-subject_2020}. This contribution naturally leads us to leverage the MD network as a prism for studying the mathematical reasoning task of models.

\textbf{Functional Localizer} \citep{fedorenko_new_2010} commonly used to identify cognitive networks rely on contrasting brain activity between two conditions. The positive condition refers to the task that robustly engages the cognitive network of interest, while the negative condition serves as a control condition that minimizes or omits such engagement. This distinction is essential for isolating signals specific to the task of interest, as the subtraction of overlapping or redundant neural activity common to both conditions preserves only the significant and relevant activation. By focusing on the differential activity between these positive and negative tasks, the functional localizer by contrast provides a robust and generalizable framework for mapping specialized cognitive networks across individuals.

\section{Methods}

In this section, we present our approach for identifying and characterizing the causal units related to ToM and MD tasks in large-scale models. Figure \ref{fig:fig1}a outlines the primary steps of the workflow.

\textbf{Models.} As an input, we consider either LLMs or VLMs. Imported from Huggingface \citep{wolf_huggingfaces_2020}, we focused on instructed-tuned versions and we selected 11 LLMs, all among the most downloaded and performant in online benchmarks. These include Llama-3.1-\{8, 70\}B, Llama-3.2-11B-Vision \citep{Dubey2024TheL3}, Qwen2.5-\{3, 7, 14, 32, 72\}B-Instruct \citep{qwen_qwen25_2025}, Mistral-Nemo-Instruct-2407, Mistral-Small-24B-Instruct-2501, and phi-3.5-mini-instruct \citep{abdin_phi-3_2024}. Additionally, for vision-language tasks, we considered Llama-3.2-\{11, 90\}B-Instruct \citep{Dubey2024TheL3} and Qwen2.5-VL-\{3, 7, 72\}B-Instruct \citep{bai_qwen25-vl_2025} models. Overall, our selection spans a wide range of parameter sizes, from 3 to 90 billions.

\textbf{Localizer Type \& Contrast Localizer.} ToM localizer distinguishes cognitive engagement by contrasting 10 False-Belief (Positive) and 10 False-Photograph (Negative) stories \citep{dodell-feder_fmri_2011}. Since MD‐system engagement scales with task difficulty, we define the MD localizer by contrasting hard-arithmetic (Positive) versus easy-arithmetic (Negative) problems (see more details in Appendix \ref{app:loc_type}). For each cognitive task, the positive and negative stimulus will be passed independently through the Language Model in a prompt format. For each stimulus, the output activation units are extracted from each transformer block (see more details in Appendix \ref{app:contrast_loc}). The \textit{top} condition refers to the units that exhibit the highest activation during the tasks. The \textit{bottom} condition serves as a control and includes units with the lowest activation for a given task, while the \textit{random} condition consists of units randomly selected from the model; this random selection was repeated 15 times for each model to ensure robustness.

\textbf{Lesioned Model \& Benchmarks.} Once we have identified the units to lesion for each condition, we lesion them by setting their activations to zero and then assess the model on benchmarks in a multiple-choice format. For ToM, we consider three unimodal datasets that are stories-based tasks, which consisted of a narrative, question and two candidate options. The three benchmarks assess false-belief ability and the narrative of these datasets are as follows: ToMi has a short narrative \citep{le_revisiting_2019, sap_neural_2022}; OpenToM has a complex narrative \citep{xu_opentom_2024}; FanToM has a dialogue based narrative between several characters \citep{kim_fantom_2023}. For MD tasks, we include the unimodal MATH dataset \citep{hendrycks_measuring_2021, zhang_multiple-choice_2024} and two multimodal datasets: MathVista \citep{lu_mathvista_2024} and MMStar \citep{chen_are_2024}. These benchmarks pose multiple choice questions on mathematical topics such as algebra, geometry, and statistics. See details in Appendix \ref{app:benchmark}.

\textbf{Score Assessment.} In contrast to \citet{alkhamissi_llm_2024}, where negative log-logits were used to select the best candidate, we instead adopt a generated-token approach—specifically chosen to assess the model’s ability to comprehend the prompt and accurately select among the candidate options. The accuracy score is computed by subtracting the baseline accuracy from the lesioned accuracy.

\section{Results}

\begin{figure}[t]
    \centering
    \begin{subfigure}[b]{1\textwidth}
        \centering
        \includegraphics[width=\textwidth]{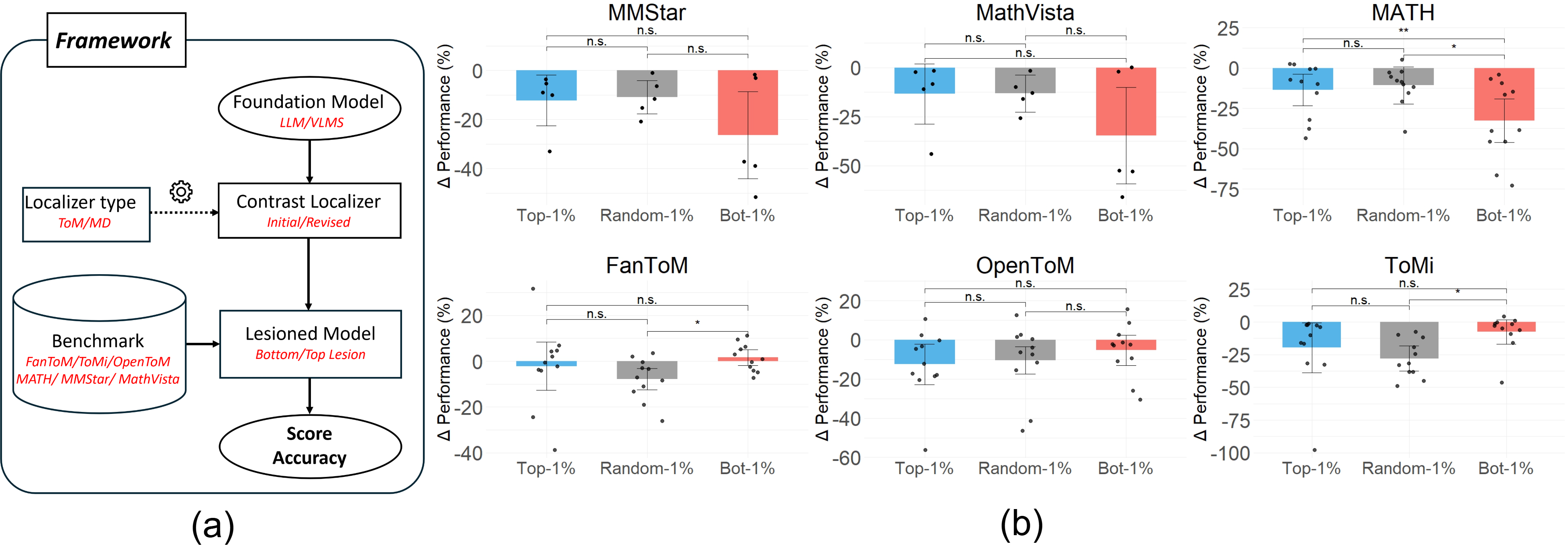}
    \end{subfigure}
    \caption{\textbf{(a) Framework workflow} outlines the general process for identifying causal units that underpin ToM or MD task in language models. The items highlighted in red represent the various options available at each step (e.g., different localizer types, benchmarks, and lesion methods).
    \textbf{(b) Comparing the performance drop by lesioning 1\% of Top, Bottom, Random units.} In each barplot, the black dots indicate the accuracy change—of the 5 VLMs for MathVista \& MMStar and 11 LLMs for the others. \textit{n.s.}: not significant; * $p<0.05$; ** $p<0.01$.}
    \label{fig:fig1}
\end{figure}

\textbf{Lesion Assessment.} 
For each model, Top and Bottom units are identified using ToM and MD localizers. Random units are shared across both tasks. Each model's performance is then evaluated on benchmarks aligned with the localizer domain: ToM localizer assesses FanToM, OpenToM, and ToMi; MD localizer assesses MMStar, MathVista, and MATH. To test generalizability, paired t-tests compare performance across three conditions: Top vs. Random and Top vs. Bottom. A significant performance drop when Top units are lesioned indicates their causal role in ToM or MD task performance. Results are shown in Figure \ref{fig:fig1}b. Across the benchmarks, The Top condition does not induce a significant performance drop compared to the control conditions across datasets. Furthermore, lesioning the Bottom units also affects model performance, which is unexpected given that these units are the least activated for the tasks. In particular, for the MATH dataset, the Bottom condition leads to a significantly larger performance drop compared to the Top and Random conditions. These findings challenge our initial assumption regarding the reliability of the contrast localizer method in accurately distinguishing units related to ToM and MD tasks.

\begin{figure}[t]
    \centering
    \begin{subfigure}[b]{0.3\textwidth}
        \includegraphics[width=\textwidth]{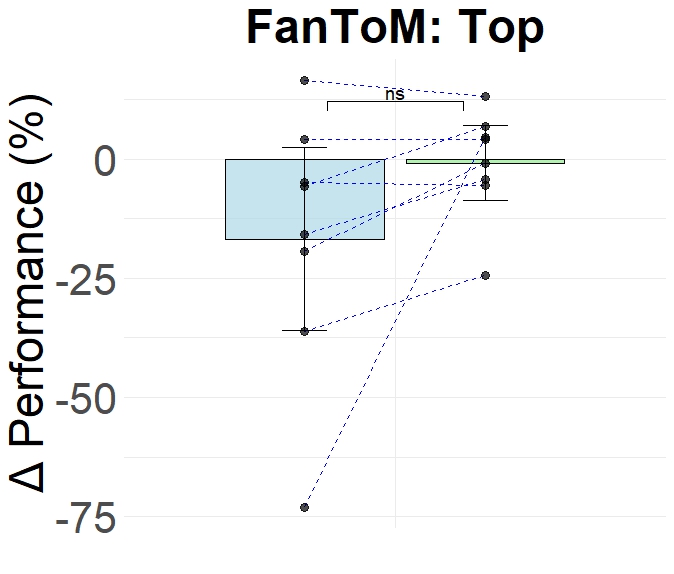}
    \end{subfigure}%
    \hspace{0.015\textwidth}%
    \begin{subfigure}[b]{0.3\textwidth}
        \includegraphics[width=\textwidth]{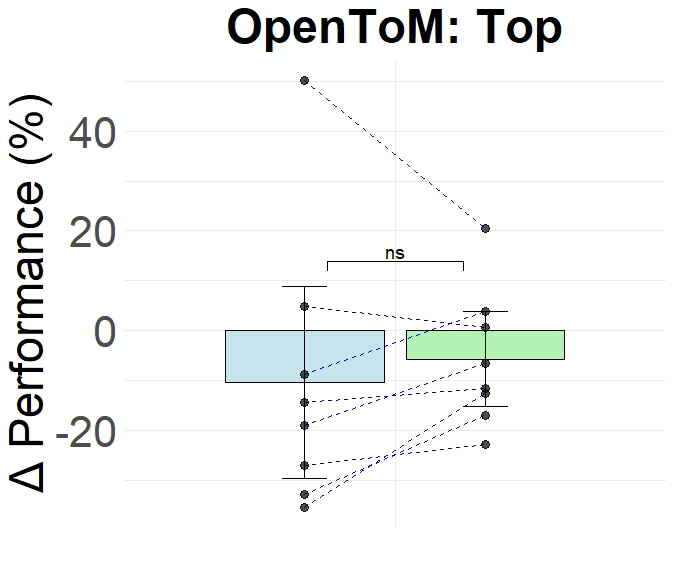}
    \end{subfigure}
    \hspace{0.015\textwidth}%
    \begin{subfigure}[b]{0.3 \textwidth}
        \includegraphics[width=\textwidth]{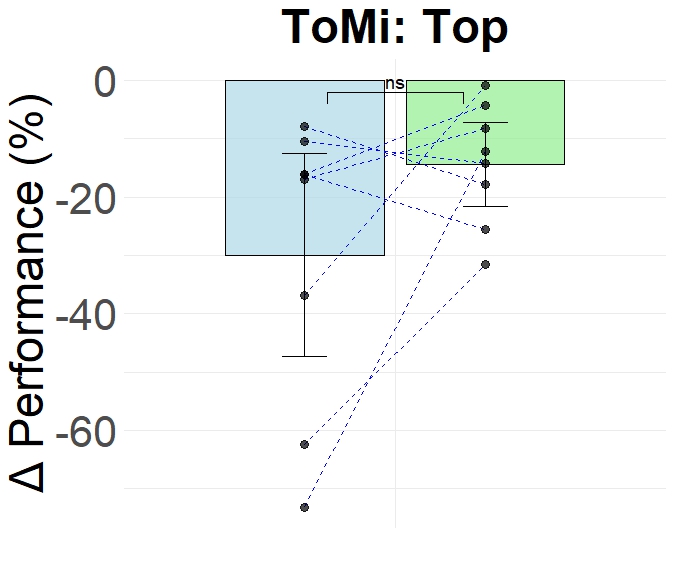}
    \end{subfigure}%
    
    \caption{\textbf{Cross-task analysis on ToM tasks.}
    The three plots compare distribution score across the three ToM datasets for 1\% lesioning on Top condition. Paired data points for each model—shown as black dots and linked by dashed lines—represent the performance differences between the MD and ToM localizers, and a paired t-test is performed. \textit{n.s.}: not significant; * $p<0.05$; ** $p<0.01$}
    \label{fig:cross-task}
\end{figure}

\textbf{Cross-task Analysis.}
To further investigate the task-specificity of the method, we performed a cross-task analysis where MD localizer is used to assess ToM benchmarks. In Figure \ref{fig:cross-task}, we report the performance impact of lesioning top-ranked units identified by both ToM and MD localizers across the ToM tasks. Interestingly, for each dataset, lesioning the MD-identified units leads to a larger performance drop than lesioning the ToM-identified units, raising additional concerns on the ability to differentiate those two cognitive tasks.

\section{Discussion}

This research aims to extend the contrast localizer design to ToM and MD tasks. The analysis on these cognitive tasks has raised many questions on the validity of the method. While the results here do not causally differentiate between ToM- and MD-relevant units, these results highlight key directions for underlying the complexity of mapping high-level cognitive functions onto LLMs\footnote{Code available on GitHub: \url{https://github.com/YassineJamaa/ToM-LargeModel}.}.

\textbf{Extending the localizer stimulus sets.} To refine the identification of sensitive activation units, it is essential to broaden both the stimulus sets and the aspects of cognitive processing they encompass for MD and ToM tasks. For the MD localizer, the current contrast is limited to comparing easy versus hard arithmetic operations (addition and subtraction). Enhancing this design by incorporating additional operations, such as division and multiplication, alongside tasks that tap into other dimensions of MD reasoning—such as logic puzzles and memory challenges—could help differentiate activation units more effectively \citep{duncan_common_2000, fedorenko_broad_2013, shashidhara_individual-subject_2020}. Similarly, the ToM localizer could be improved by broadening its stimulus array which would provide richer contrast. Another method, described by \citep{bruneau_distinct_2012}, employs an extensive stimulus set that contrasts 48 stories depicting varying degrees of physical or emotional pain with 48 corresponding versions in which these painful elements have been removed. By applying the both contrasts and identifying the common activation units across them, one can define a robust positive set that more reliably reflects ToM units while capturing a broader spectrum of ToM.

\textbf{Granularity of Extracted Units.} In our study, we focused exclusively on the output activation units from each transformer block, as these units directly influence the generation of coherent tokens as shown in \citet{alkhamissi_llm_2024}. However, concentrating solely on these output activations raises the question of whether this level of granularity is sufficient for dissecting complex reasoning tasks, such as mathematical and social reasoning. These tasks likely rely on a broader array of internal computations beyond just the final outputs. A more comprehensive approach would involve examining intermediate representations and hidden activations within each transformer block. By incorporating these additional activation units into the analysis, we can gain deeper insights into how reasoning processes are distributed throughout the model’s architecture and better understand the neural mechanisms underlying advanced cognitive functions.

\clearpage
\bibliography{colm2025_conference}

\begin{thebibliography}{43}
\providecommand{\natexlab}[1]{#1}
\providecommand{\url}[1]{\texttt{#1}}
\expandafter\ifx\csname urlstyle\endcsname\relax
  \providecommand{\doi}[1]{doi: #1}\else
  \providecommand{\doi}{doi: \begingroup \urlstyle{rm}\Url}\fi

\bibitem[Abdin et~al.(2024)Abdin, Aneja, Awadalla, and Awadallah]{abdin_phi-3_2024}
Marah Abdin, Jyoti Aneja, Hany Awadalla, and Awadallah.
\newblock Phi-3 {Technical} {Report}: {A} {Highly} {Capable} {Language} {Model} {Locally} on {Your} {Phone}, August 2024.
\newblock URL \url{http://arxiv.org/abs/2404.14219}.
\newblock arXiv:2404.14219 [cs].

\bibitem[AlKhamissi et~al.(2024)AlKhamissi, Tuckute, Bosselut, and Schrimpf]{alkhamissi_llm_2024}
Badr AlKhamissi, Greta Tuckute, Antoine Bosselut, and Martin Schrimpf.
\newblock The {LLM} {Language} {Network}: {A} {Neuroscientific} {Approach} for {Identifying} {Causally} {Task}-{Relevant} {Units}.
\newblock 2024.

\bibitem[Assem et~al.(2020)Assem, Glasser, Van~Essen, and Duncan]{assem_domain-general_2020}
Moataz Assem, Matthew~F Glasser, David~C Van~Essen, and John Duncan.
\newblock A {Domain}-{General} {Cognitive} {Core} {Defined} in {Multimodally} {Parcellated} {Human} {Cortex}.
\newblock \emph{Cerebral Cortex}, 30\penalty0 (8):\penalty0 4361--4380, June 2020.
\newblock ISSN 1047-3211, 1460-2199.
\newblock \doi{10.1093/cercor/bhaa023}.
\newblock URL \url{https://academic.oup.com/cercor/article/30/8/4361/5815289}.

\bibitem[Bai et~al.(2025)Bai, Chen, Liu, Wang, Ge, Song, Dang, Wang, Wang, Tang, Zhong, Zhu, Yang, Li, Wan, Wang, Ding, Fu, Xu, Ye, Zhang, Xie, Cheng, Zhang, Yang, Xu, and Lin]{bai_qwen25-vl_2025}
Shuai Bai, Keqin Chen, Xuejing Liu, Jialin Wang, Wenbin Ge, Sibo Song, Kai Dang, Peng Wang, Shijie Wang, Jun Tang, Humen Zhong, Yuanzhi Zhu, Mingkun Yang, Zhaohai Li, Jianqiang Wan, Pengfei Wang, Wei Ding, Zheren Fu, Yiheng Xu, Jiabo Ye, Xi~Zhang, Tianbao Xie, Zesen Cheng, Hang Zhang, Zhibo Yang, Haiyang Xu, and Junyang Lin.
\newblock Qwen2.5-{VL} {Technical} {Report}, February 2025.
\newblock URL \url{http://arxiv.org/abs/2502.13923}.
\newblock arXiv:2502.13923 [cs].

\bibitem[Baker et~al.(2009)Baker, Saxe, and Tenenbaum]{baker_action_2009}
Chris~L. Baker, Rebecca Saxe, and Joshua~B. Tenenbaum.
\newblock Action understanding as inverse planning.
\newblock \emph{Cognition}, 113\penalty0 (3):\penalty0 329--349, December 2009.
\newblock ISSN 00100277.
\newblock \doi{10.1016/j.cognition.2009.07.005}.
\newblock URL \url{https://linkinghub.elsevier.com/retrieve/pii/S0010027709001607}.

\bibitem[Baker et~al.(2017)Baker, Jara-Ettinger, Saxe, and Tenenbaum]{baker_rational_2017}
Chris~L. Baker, Julian Jara-Ettinger, Rebecca Saxe, and Joshua~B. Tenenbaum.
\newblock Rational quantitative attribution of beliefs, desires and percepts in human mentalizing.
\newblock \emph{Nature Human Behaviour}, 1\penalty0 (4):\penalty0 0064, March 2017.
\newblock ISSN 2397-3374.
\newblock \doi{10.1038/s41562-017-0064}.
\newblock URL \url{https://www.nature.com/articles/s41562-017-0064}.

\bibitem[Baron-Cohen et~al.(1985)Baron-Cohen, Leslie, and Frith]{baron-cohen_does_1985}
Simon Baron-Cohen, Alan~M. Leslie, and Uta Frith.
\newblock Does the autistic child have a “theory of mind” ?
\newblock \emph{Cognition}, 21\penalty0 (1):\penalty0 37--46, October 1985.
\newblock ISSN 00100277.
\newblock \doi{10.1016/0010-0277(85)90022-8}.
\newblock URL \url{https://linkinghub.elsevier.com/retrieve/pii/0010027785900228}.

\bibitem[Bischetti et~al.(2023)Bischetti, Ceccato, Lecce, Cavallini, and Bambini]{bischetti_pragmatics_2023}
Luca Bischetti, Irene Ceccato, Serena Lecce, Elena Cavallini, and Valentina Bambini.
\newblock Pragmatics and theory of mind in older adults’ humor comprehension.
\newblock \emph{Current Psychology}, 42\penalty0 (19):\penalty0 16191--16207, July 2023.
\newblock ISSN 1046-1310, 1936-4733.
\newblock \doi{10.1007/s12144-019-00295-w}.
\newblock URL \url{https://link.springer.com/10.1007/s12144-019-00295-w}.

\bibitem[Bruneau et~al.(2012)Bruneau, Pluta, and Saxe]{bruneau_distinct_2012}
Emile~G. Bruneau, Agnieszka Pluta, and Rebecca Saxe.
\newblock Distinct roles of the ‘{Shared} {Pain}’ and ‘{Theory} of {Mind}’ networks in processing others’ emotional suffering.
\newblock \emph{Neuropsychologia}, 50\penalty0 (2):\penalty0 219--231, January 2012.
\newblock ISSN 00283932.
\newblock \doi{10.1016/j.neuropsychologia.2011.11.008}.
\newblock URL \url{https://linkinghub.elsevier.com/retrieve/pii/S0028393211005082}.

\bibitem[Chen et~al.(2024)Chen, Li, Dong, Zhang, Zang, Chen, Duan, Wang, Qiao, Lin, and Zhao]{chen_are_2024}
Lin Chen, Jinsong Li, Xiaoyi Dong, Pan Zhang, Yuhang Zang, Zehui Chen, Haodong Duan, Jiaqi Wang, Yu~Qiao, Dahua Lin, and Feng Zhao.
\newblock Are {We} on the {Right} {Way} for {Evaluating} {Large} {Vision}-{Language} {Models}?, April 2024.
\newblock URL \url{http://arxiv.org/abs/2403.20330}.
\newblock arXiv:2403.20330 [cs].

\bibitem[Cole \& Schneider(2007)Cole and Schneider]{cole_cognitive_2007}
Michael~W. Cole and Walter Schneider.
\newblock The cognitive control network: {Integrated} cortical regions with dissociable functions.
\newblock \emph{NeuroImage}, 37\penalty0 (1):\penalty0 343--360, August 2007.
\newblock ISSN 10538119.
\newblock \doi{10.1016/j.neuroimage.2007.03.071}.
\newblock URL \url{https://linkinghub.elsevier.com/retrieve/pii/S1053811907002820}.

\bibitem[Dodell-Feder et~al.(2011)Dodell-Feder, Koster-Hale, Bedny, and Saxe]{dodell-feder_fmri_2011}
David Dodell-Feder, Jorie Koster-Hale, Marina Bedny, and Rebecca Saxe.
\newblock {fMRI} item analysis in a theory of mind task.
\newblock \emph{NeuroImage}, 55\penalty0 (2):\penalty0 705--712, March 2011.
\newblock ISSN 10538119.
\newblock \doi{10.1016/j.neuroimage.2010.12.040}.
\newblock URL \url{https://linkinghub.elsevier.com/retrieve/pii/S1053811910016241}.

\bibitem[Dubey et~al.(2024)Dubey, Jauhri, and Pandey]{Dubey2024TheL3}
Abhimanyu Dubey, Abhinav Jauhri, and Abhinav Pandey.
\newblock The llama 3 herd of models.
\newblock \emph{ArXiv}, abs/2407.21783, 2024.
\newblock URL \url{https://api.semanticscholar.org/CorpusID:271571434}.

\bibitem[Duncan \& Owen(2000)Duncan and Owen]{duncan_common_2000}
John Duncan and Adrian~M Owen.
\newblock Common regions of the human frontal lobe recruited by diverse cognitive demands.
\newblock \emph{Trends in Neurosciences}, 23\penalty0 (10):\penalty0 475--483, October 2000.
\newblock ISSN 01662236.
\newblock \doi{10.1016/S0166-2236(00)01633-7}.
\newblock URL \url{https://linkinghub.elsevier.com/retrieve/pii/S0166223600016337}.

\bibitem[Fedorenko et~al.(2010)Fedorenko, Hsieh, Nieto-Castañón, Whitfield-Gabrieli, and Kanwisher]{fedorenko_new_2010}
Evelina Fedorenko, Po-Jang Hsieh, Alfonso Nieto-Castañón, Susan Whitfield-Gabrieli, and Nancy Kanwisher.
\newblock New {Method} for {fMRI} {Investigations} of {Language}: {Defining} {ROIs} {Functionally} in {Individual} {Subjects}.
\newblock \emph{Journal of Neurophysiology}, 104\penalty0 (2):\penalty0 1177--1194, August 2010.
\newblock ISSN 0022-3077, 1522-1598.
\newblock \doi{10.1152/jn.00032.2010}.
\newblock URL \url{https://www.physiology.org/doi/10.1152/jn.00032.2010}.

\bibitem[Fedorenko et~al.(2013)Fedorenko, Duncan, and Kanwisher]{fedorenko_broad_2013}
Evelina Fedorenko, John Duncan, and Nancy Kanwisher.
\newblock Broad domain generality in focal regions of frontal and parietal cortex.
\newblock \emph{Proceedings of the National Academy of Sciences}, 110\penalty0 (41):\penalty0 16616--16621, October 2013.
\newblock ISSN 0027-8424, 1091-6490.
\newblock \doi{10.1073/pnas.1315235110}.
\newblock URL \url{https://pnas.org/doi/full/10.1073/pnas.1315235110}.

\bibitem[Gallagher et~al.(2000)Gallagher, Happé, Brunswick, Fletcher, Frith, and Frith]{gallagher_reading_2000}
H.L Gallagher, F~Happé, N~Brunswick, P.C Fletcher, U~Frith, and C.D Frith.
\newblock Reading the mind in cartoons and stories: an {fMRI} study of ‘theory of mind’ in verbal and nonverbal tasks.
\newblock \emph{Neuropsychologia}, 38\penalty0 (1):\penalty0 11--21, January 2000.
\newblock ISSN 00283932.
\newblock \doi{10.1016/S0028-3932(99)00053-6}.
\newblock URL \url{https://linkinghub.elsevier.com/retrieve/pii/S0028393299000536}.

\bibitem[Geiger et~al.(2021)Geiger, Lu, Icard, and Potts]{geiger_causal_2021}
Atticus Geiger, Hanson Lu, Thomas Icard, and Christopher Potts.
\newblock Causal {Abstractions} of {Neural} {Networks}, October 2021.
\newblock URL \url{http://arxiv.org/abs/2106.02997}.
\newblock arXiv:2106.02997 [cs].

\bibitem[Giadikiaroglou et~al.(2024)Giadikiaroglou, Lymperaiou, Filandrianos, and Stamou]{giadikiaroglou_puzzle_2024}
Panagiotis Giadikiaroglou, Maria Lymperaiou, Giorgos Filandrianos, and Giorgos Stamou.
\newblock Puzzle {Solving} using {Reasoning} of {Large} {Language} {Models}: {A} {Survey}, September 2024.
\newblock URL \url{http://arxiv.org/abs/2402.11291}.
\newblock arXiv:2402.11291 [cs].

\bibitem[Hendrycks et~al.(2021)Hendrycks, Burns, Kadavath, Arora, Basart, Tang, Song, and Steinhardt]{hendrycks_measuring_2021}
Dan Hendrycks, Collin Burns, Saurav Kadavath, Akul Arora, Steven Basart, Eric Tang, Dawn Song, and Jacob Steinhardt.
\newblock Measuring {Mathematical} {Problem} {Solving} {With} the {MATH} {Dataset}, November 2021.
\newblock URL \url{http://arxiv.org/abs/2103.03874}.
\newblock arXiv:2103.03874 [cs].

\bibitem[Hsu \& Cheung(2013)Hsu and Cheung]{hsu_two_2013}
Yik~Kwan Hsu and Him Cheung.
\newblock Two mentalizing capacities and the understanding of two types of lie telling in children.
\newblock \emph{Developmental Psychology}, 49\penalty0 (9):\penalty0 1650--1659, September 2013.
\newblock ISSN 1939-0599, 0012-1649.
\newblock \doi{10.1037/a0031128}.
\newblock URL \url{https://doi.apa.org/doi/10.1037/a0031128}.

\bibitem[Kim et~al.(2023)Kim, Sclar, Zhou, Bras, Kim, Choi, and Sap]{kim_fantom_2023}
Hyunwoo Kim, Melanie Sclar, Xuhui Zhou, Ronan~Le Bras, Gunhee Kim, Yejin Choi, and Maarten Sap.
\newblock {FANToM}: {A} {Benchmark} for {Stress}-testing {Machine} {Theory} of {Mind} in {Interactions}, October 2023.
\newblock URL \url{http://arxiv.org/abs/2310.15421}.
\newblock arXiv:2310.15421 [cs].

\bibitem[Le et~al.(2019)Le, Boureau, and Nickel]{le_revisiting_2019}
Matthew Le, Y-Lan Boureau, and Maximilian Nickel.
\newblock Revisiting the {Evaluation} of {Theory} of {Mind} through {Question} {Answering}.
\newblock In \emph{Proceedings of the 2019 {Conference} on {Empirical} {Methods} in {Natural} {Language} {Processing} and the 9th {International} {Joint} {Conference} on {Natural} {Language} {Processing} ({EMNLP}-{IJCNLP})}, pp.\  5871--5876, Hong Kong, China, 2019. Association for Computational Linguistics.
\newblock \doi{10.18653/v1/D19-1598}.
\newblock URL \url{https://www.aclweb.org/anthology/D19-1598}.

\bibitem[Leslie et~al.(2006)Leslie, Knobe, and Cohen]{leslie_acting_2006}
Alan~M. Leslie, Joshua Knobe, and Adam Cohen.
\newblock Acting {Intentionally} and the {Side}-{Effect} {Effect}: {Theory} of {Mind} and {Moral} {Judgment}.
\newblock \emph{Psychological Science}, 17\penalty0 (5):\penalty0 421--427, May 2006.
\newblock ISSN 0956-7976, 1467-9280.
\newblock \doi{10.1111/j.1467-9280.2006.01722.x}.
\newblock URL \url{https://journals.sagepub.com/doi/10.1111/j.1467-9280.2006.01722.x}.

\bibitem[Lu et~al.(2024)Lu, Bansal, Xia, Liu, Li, Hajishirzi, Cheng, Chang, Galley, and Gao]{lu_mathvista_2024}
Pan Lu, Hritik Bansal, Tony Xia, Jiacheng Liu, Chunyuan Li, Hannaneh Hajishirzi, Hao Cheng, Kai-Wei Chang, Michel Galley, and Jianfeng Gao.
\newblock {MathVista}: {Evaluating} {Mathematical} {Reasoning} of {Foundation} {Models} in {Visual} {Contexts}, January 2024.
\newblock URL \url{http://arxiv.org/abs/2310.02255}.
\newblock arXiv:2310.02255 [cs].

\bibitem[Qwen et~al.(2025)Qwen, Yang, Yang, Zhang, Hui, Zheng, Yu, Li, Liu, Huang, Wei, Lin, Yang, Tu, Zhang, Yang, Yang, Zhou, Lin, Dang, Lu, Bao, Yang, Yu, Li, Xue, Zhang, Zhu, Men, Lin, Li, Tang, Xia, Ren, Ren, Fan, Su, Zhang, Wan, Liu, Cui, Zhang, and Qiu]{qwen_qwen25_2025}
Qwen, An~Yang, Baosong Yang, Beichen Zhang, Binyuan Hui, Bo~Zheng, Bowen Yu, Chengyuan Li, Dayiheng Liu, Fei Huang, Haoran Wei, Huan Lin, Jian Yang, Jianhong Tu, Jianwei Zhang, Jianxin Yang, Jiaxi Yang, Jingren Zhou, Junyang Lin, Kai Dang, Keming Lu, Keqin Bao, Kexin Yang, Le~Yu, Mei Li, Mingfeng Xue, Pei Zhang, Qin Zhu, Rui Men, Runji Lin, Tianhao Li, Tianyi Tang, Tingyu Xia, Xingzhang Ren, Xuancheng Ren, Yang Fan, Yang Su, Yichang Zhang, Yu~Wan, Yuqiong Liu, Zeyu Cui, Zhenru Zhang, and Zihan Qiu.
\newblock Qwen2.5 {Technical} {Report}, January 2025.
\newblock URL \url{http://arxiv.org/abs/2412.15115}.
\newblock arXiv:2412.15115 [cs].

\bibitem[Sap et~al.(2022)Sap, Le~Bras, Fried, and Choi]{sap_neural_2022}
Maarten Sap, Ronan Le~Bras, Daniel Fried, and Yejin Choi.
\newblock Neural {Theory}-of-{Mind}? {On} the {Limits} of {Social} {Intelligence} in {Large} {LMs}.
\newblock In \emph{Proceedings of the 2022 {Conference} on {Empirical} {Methods} in {Natural} {Language} {Processing}}, pp.\  3762--3780, Abu Dhabi, United Arab Emirates, 2022. Association for Computational Linguistics.
\newblock \doi{10.18653/v1/2022.emnlp-main.248}.
\newblock URL \url{https://aclanthology.org/2022.emnlp-main.248}.

\bibitem[Saxe \& Powell(2006)Saxe and Powell]{saxe_its_2006}
Rebecca Saxe and Lindsey~J. Powell.
\newblock It's the {Thought} {That} {Counts}: {Specific} {Brain} {Regions} for {One} {Component} of {Theory} of {Mind}.
\newblock \emph{Psychological Science}, 17\penalty0 (8):\penalty0 692--699, August 2006.
\newblock ISSN 0956-7976, 1467-9280.
\newblock \doi{10.1111/j.1467-9280.2006.01768.x}.
\newblock URL \url{https://journals.sagepub.com/doi/10.1111/j.1467-9280.2006.01768.x}.

\bibitem[Schrimpf et~al.(2018)Schrimpf, Kubilius, Hong, Majaj, Rajalingham, Issa, Kar, Bashivan, Prescott-Roy, Geiger, Schmidt, Yamins, and DiCarlo]{schrimpf_brain-score_2018}
Martin Schrimpf, Jonas Kubilius, Ha~Hong, Najib~J. Majaj, Rishi Rajalingham, Elias~B. Issa, Kohitij Kar, Pouya Bashivan, Jonathan Prescott-Roy, Franziska Geiger, Kailyn Schmidt, Daniel L.~K. Yamins, and James~J. DiCarlo.
\newblock Brain-{Score}: {Which} {Artificial} {Neural} {Network} for {Object} {Recognition} is most {Brain}-{Like}?, September 2018.
\newblock URL \url{http://biorxiv.org/lookup/doi/10.1101/407007}.

\bibitem[Schrimpf et~al.(2021)Schrimpf, Blank, Tuckute, Kauf, Hosseini, Kanwisher, Tenenbaum, and Fedorenko]{schrimpf_neural_2021}
Martin Schrimpf, Idan~Asher Blank, Greta Tuckute, Carina Kauf, Eghbal~A. Hosseini, Nancy Kanwisher, Joshua~B. Tenenbaum, and Evelina Fedorenko.
\newblock The neural architecture of language: {Integrative} modeling converges on predictive processing.
\newblock \emph{Proceedings of the National Academy of Sciences}, 118\penalty0 (45):\penalty0 e2105646118, November 2021.
\newblock ISSN 0027-8424, 1091-6490.
\newblock \doi{10.1073/pnas.2105646118}.
\newblock URL \url{https://pnas.org/doi/full/10.1073/pnas.2105646118}.

\bibitem[Schrimpf et~al.(2024)Schrimpf, McGrath, Margalit, and DiCarlo]{schrimpf_topographic_2024}
Martin Schrimpf, Paul McGrath, Eshed Margalit, and James~J. DiCarlo.
\newblock Do {Topographic} {Deep} {ANN} {Models} of the {Primate} {Ventral} {Stream} {Predict} the {Perceptual} {Effects} of {Direct} {IT} {Cortical} {Interventions}?, January 2024.
\newblock URL \url{http://biorxiv.org/lookup/doi/10.1101/2024.01.09.572970}.

\bibitem[Sclar et~al.(2023)Sclar, Kumar, West, Suhr, Choi, and Tsvetkov]{sclar_minding_2023}
Melanie Sclar, Sachin Kumar, Peter West, Alane Suhr, Yejin Choi, and Yulia Tsvetkov.
\newblock Minding {Language} {Models}’ ({Lack} of) {Theory} of {Mind}: {A} {Plug}-and-{Play} {Multi}-{Character} {Belief} {Tracker}.
\newblock In \emph{Proceedings of the 61st {Annual} {Meeting} of the {Association} for {Computational} {Linguistics} ({Volume} 1: {Long} {Papers})}, pp.\  13960--13980, Toronto, Canada, 2023. Association for Computational Linguistics.
\newblock \doi{10.18653/v1/2023.acl-long.780}.
\newblock URL \url{https://aclanthology.org/2023.acl-long.780}.

\bibitem[Shashidhara et~al.(2020)Shashidhara, Spronkers, and Erez]{shashidhara_individual-subject_2020}
Sneha Shashidhara, Floortje~S. Spronkers, and Yaara Erez.
\newblock Individual-subject {Functional} {Localization} {Increases} {Univariate} {Activation} but {Not} {Multivariate} {Pattern} {Discriminability} in the “{Multiple}-demand” {Frontoparietal} {Network}.
\newblock \emph{Journal of Cognitive Neuroscience}, 32\penalty0 (7):\penalty0 1348--1368, July 2020.
\newblock ISSN 0898-929X, 1530-8898.
\newblock \doi{10.1162/jocn_a_01554}.
\newblock URL \url{https://direct.mit.edu/jocn/article/32/7/1348/95439/Individual-subject-Functional-Localization}.

\bibitem[Sosa et~al.()Sosa, Ullman, Tenenbaum, Gershman, and Gerstenberg]{sosa_moral_nodate}
Felix~A Sosa, Tomer Ullman, Joshua~B Tenenbaum, Samuel~J Gershman, and Tobias Gerstenberg.
\newblock Moral dynamics: {Grounding} moral judgment in intuitive physics and intuitive psychology.

\bibitem[Spotorno et~al.(2012)Spotorno, Koun, Prado, Van Der~Henst, and Noveck]{spotorno_neural_2012}
Nicola Spotorno, Eric Koun, Jérôme Prado, Jean-Baptiste Van Der~Henst, and Ira~A. Noveck.
\newblock Neural evidence that utterance-processing entails mentalizing: {The} case of irony.
\newblock \emph{NeuroImage}, 63\penalty0 (1):\penalty0 25--39, October 2012.
\newblock ISSN 10538119.
\newblock \doi{10.1016/j.neuroimage.2012.06.046}.
\newblock URL \url{https://linkinghub.elsevier.com/retrieve/pii/S1053811912006611}.

\bibitem[Street et~al.(2024)Street, Siy, Keeling, Baranes, Barnett, McKibben, Kanyere, Lentz, Arcas, and Dunbar]{street_llms_2024}
Winnie Street, John~Oliver Siy, Geoff Keeling, Adrien Baranes, Benjamin Barnett, Michael McKibben, Tatenda Kanyere, Alison Lentz, Blaise Aguera~y Arcas, and Robin I.~M. Dunbar.
\newblock {LLMs} achieve adult human performance on higher-order theory of mind tasks, May 2024.
\newblock URL \url{http://arxiv.org/abs/2405.18870}.
\newblock arXiv:2405.18870 [cs].

\bibitem[Sun et~al.(2024)Sun, Zheng, Xie, Liu, Chu, Qiu, Xu, Ding, Li, Geng, Wu, Wang, Chen, Yin, Ren, Fu, He, Yuan, Liu, Liu, Li, Dong, Cheng, Zhang, Heng, Dai, Luo, Wang, Wen, Qiu, Guo, Xiong, Liu, and Li]{sun_survey_2024}
Jiankai Sun, Chuanyang Zheng, Enze Xie, Zhengying Liu, Ruihang Chu, Jianing Qiu, Jiaqi Xu, Mingyu Ding, Hongyang Li, Mengzhe Geng, Yue Wu, Wenhai Wang, Junsong Chen, Zhangyue Yin, Xiaozhe Ren, Jie Fu, Junxian He, Wu~Yuan, Qi~Liu, Xihui Liu, Yu~Li, Hao Dong, Yu~Cheng, Ming Zhang, Pheng~Ann Heng, Jifeng Dai, Ping Luo, Jingdong Wang, Ji-Rong Wen, Xipeng Qiu, Yike Guo, Hui Xiong, Qun Liu, and Zhenguo Li.
\newblock A {Survey} of {Reasoning} with {Foundation} {Models}, January 2024.
\newblock URL \url{http://arxiv.org/abs/2312.11562}.
\newblock arXiv:2312.11562 [cs].

\bibitem[Van~Duijn et~al.(2023)Van~Duijn, Van~Dijk, Kouwenhoven, De~Valk, Spruit, and vanderPutten]{van_duijn_theory_2023}
Max Van~Duijn, Bram Van~Dijk, Tom Kouwenhoven, Werner De~Valk, Marco Spruit, and Peter vanderPutten.
\newblock Theory of {Mind} in {Large} {Language} {Models}: {Examining} {Performance} of 11 {State}-of-the-{Art} models vs. {Children} {Aged} 7-10 on {Advanced} {Tests}.
\newblock In \emph{Proceedings of the 27th {Conference} on {Computational} {Natural} {Language} {Learning} ({CoNLL})}, pp.\  389--402, Singapore, 2023. Association for Computational Linguistics.
\newblock \doi{10.18653/v1/2023.conll-1.25}.
\newblock URL \url{https://aclanthology.org/2023.conll-1.25}.

\bibitem[Wang et~al.(2022)Wang, Variengien, Conmy, Shlegeris, and Steinhardt]{wang_interpretability_2022}
Kevin Wang, Alexandre Variengien, Arthur Conmy, Buck Shlegeris, and Jacob Steinhardt.
\newblock Interpretability in the {Wild}: a {Circuit} for {Indirect} {Object} {Identification} in {GPT}-2 small, November 2022.
\newblock URL \url{http://arxiv.org/abs/2211.00593}.
\newblock arXiv:2211.00593 [cs].

\bibitem[Wolf et~al.(2020)Wolf, Debut, Sanh, Chaumond, Delangue, Moi, Cistac, Rault, Louf, Funtowicz, Davison, Shleifer, Platen, Ma, Jernite, Plu, Xu, Scao, Gugger, Drame, Lhoest, and Rush]{wolf_huggingfaces_2020}
Thomas Wolf, Lysandre Debut, Victor Sanh, Julien Chaumond, Clement Delangue, Anthony Moi, Pierric Cistac, Tim Rault, Rémi Louf, Morgan Funtowicz, Joe Davison, Sam Shleifer, Patrick~von Platen, Clara Ma, Yacine Jernite, Julien Plu, Canwen Xu, Teven~Le Scao, Sylvain Gugger, Mariama Drame, Quentin Lhoest, and Alexander~M. Rush.
\newblock {HuggingFace}'s {Transformers}: {State}-of-the-art {Natural} {Language} {Processing}, July 2020.
\newblock URL \url{http://arxiv.org/abs/1910.03771}.
\newblock arXiv:1910.03771 [cs].

\bibitem[Woolgar et~al.(2010)Woolgar, Parr, Cusack, Thompson, Nimmo-Smith, Torralva, Roca, Antoun, Manes, and Duncan]{woolgar_fluid_2010}
Alexandra Woolgar, Alice Parr, Rhodri Cusack, Russell Thompson, Ian Nimmo-Smith, Teresa Torralva, Maria Roca, Nagui Antoun, Facundo Manes, and John Duncan.
\newblock Fluid intelligence loss linked to restricted regions of damage within frontal and parietal cortex.
\newblock \emph{Proceedings of the National Academy of Sciences}, 107\penalty0 (33):\penalty0 14899--14902, August 2010.
\newblock ISSN 0027-8424, 1091-6490.
\newblock \doi{10.1073/pnas.1007928107}.
\newblock URL \url{https://pnas.org/doi/full/10.1073/pnas.1007928107}.

\bibitem[Xu et~al.(2024)Xu, Zhao, Zhu, Du, and He]{xu_opentom_2024}
Hainiu Xu, Runcong Zhao, Lixing Zhu, Jinhua Du, and Yulan He.
\newblock {OpenToM}: {A} {Comprehensive} {Benchmark} for {Evaluating} {Theory}-of-{Mind} {Reasoning} {Capabilities} of {Large} {Language} {Models}, June 2024.
\newblock URL \url{http://arxiv.org/abs/2402.06044}.
\newblock arXiv:2402.06044 [cs].

\bibitem[Zhang et~al.(2024)Zhang, Jiang, Xu, Hao, and Wang]{zhang_multiple-choice_2024}
Ziyin Zhang, Zhaokun Jiang, Lizhen Xu, Hongkun Hao, and Rui Wang.
\newblock Multiple-{Choice} {Questions} are {Efficient} and {Robust} {LLM} {Evaluators}, June 2024.
\newblock URL \url{http://arxiv.org/abs/2405.11966}.
\newblock arXiv:2405.11966 [cs].

\end{thebibliography}
\bibliographystyle{colm2025_conference}

\appendix

\section{Contrast Localizer}
\label{app:contrast_loc}

\begin{figure}[h!]
    \centering
    \begin{subfigure}[b]{0.3\textwidth}
        \centering
        \includegraphics[width=\textwidth]{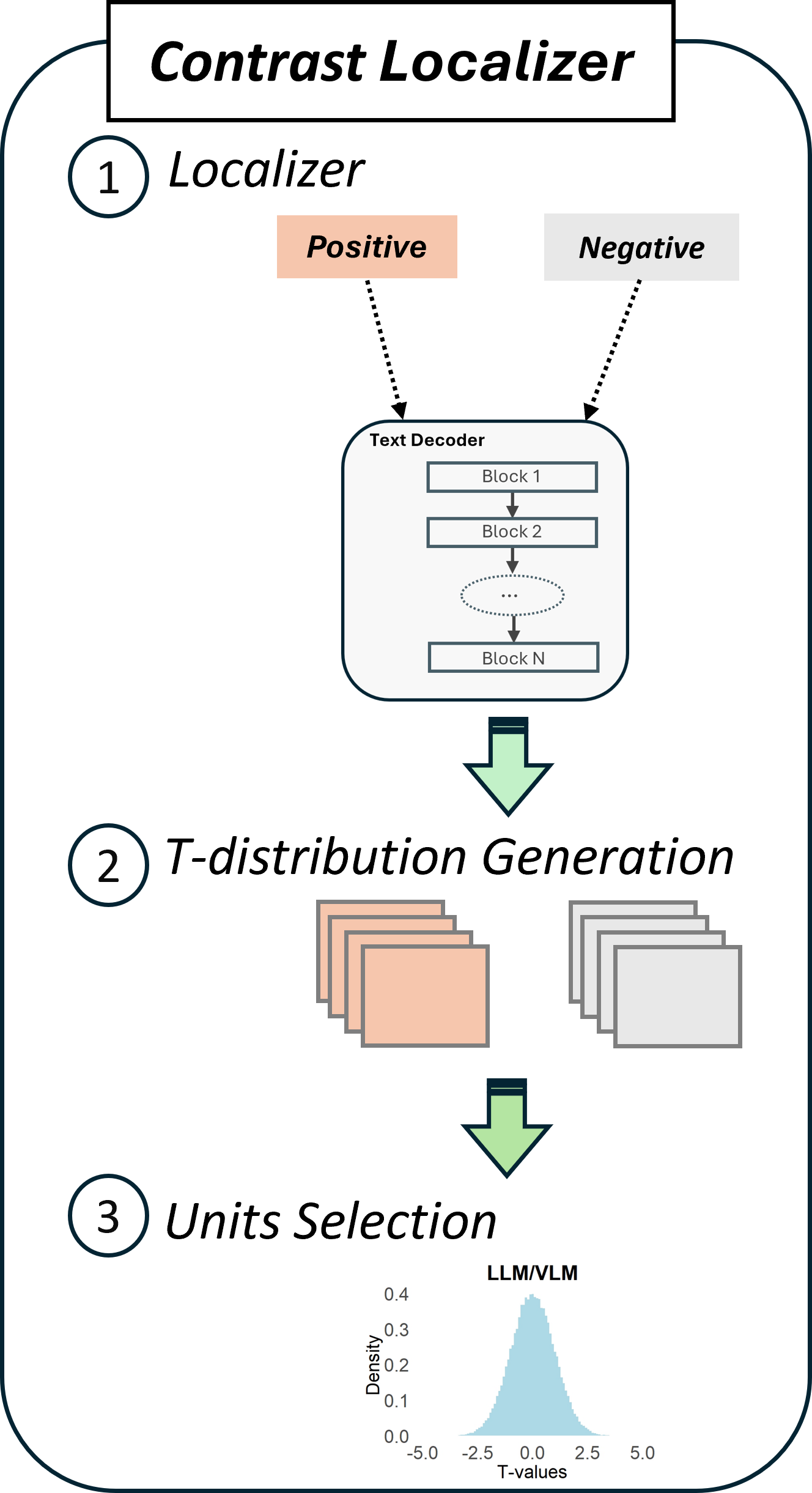}
    \end{subfigure}
    \caption{\textbf{Contrast Localizer method} zooms in on the three key decision stages for isolating specific activation units. First, a localizer contrasts positive and negative samples to pinpoint relevant regions of the model. Next, a t-distribution is generated to quantify the significance of each unit’s activation differences. Finally, a unit selection step identifies and ranks the top candidates for further causal analysis.}
    \label{contrast_localizer:scheme}
\end{figure}

Built on the approach from \citep{alkhamissi_llm_2024}, the contrast localizer is designed to characterize the distribution of activation units in response to different stimuli. Specifically, each unit is assigned a score that quantifies its relative activation level under positive versus negative stimuli. This score indicates whether a unit is preferentially activated by positive inputs, by negative inputs, or exhibits consistent activation across both conditions, thereby providing insight into the unit’s functional role within the model. The subsequent sections present a comprehensive breakdown of each step, as further illustrated by figure \ref{contrast_localizer:scheme}.

\subsection{Localizer}

The contrast localizer consists of two sets of task conditions used to identify a cognitive network. The positive contrast set, denoted as \( \mathit{P} \), consists of \( \mathit{N_p} \) prompts designed to engage the targeted cognitive process. Each prompt \( P_i \in \mathit{P} \) is indexed by \( i \), where \( i \in \{1, 2, \dots, \mathit{N_p}\} \). The negative contrast set, denoted as \( \mathit{N} \), consists of \( \mathit{N_n} \) prompts that serve as a control condition, minimizing engagement of the targeted cognitive process. Each prompt \( N_j \in \mathit{N} \) is indexed by \( j \), where \( j \in \{1, 2, \dots, \mathit{N_n}\} \). These two sets, \( P \) and \( N \), are processed independently by the Language Model, yielding output activation units from the \( M \) transformer blocks for each prompt. Specifically, for prompts in the positive contrast set, the activations are represented as \( A_i^{(P)} \in \mathbb{R}^{\mathit{M} \times \mathit{L} \times \mathit{H}} \) where \( i \in \{1, 2, \dots, \mathit{N_p}\} \), while for prompts in the negative contrast set, they are denoted as \( A_j^{(N)} \in \mathbb{R}^{\mathit{M} \times \mathit{L} \times \mathit{H}} \) where \( j \in \{1, 2, \dots, \mathit{N_n}\} \). The activation units are then averaged along the token sequence, resulting in the final representations
\[
A_i^{(P)} \in \mathbb{R}^{\mathit{M} \times \mathit{H}} \quad \text{and} \quad A_j^{(N)} \in \mathbb{R}^{\mathit{M} \times \mathit{H}},
\]
for the positive and negative contrast sets, respectively. Once the model's activation units have been extracted for both the positive and negative prompt sets, their distributions will be statistically compared, as described in the following section.

\subsection{T-distribution Generation}

Each activation unit \( u_{ij} \) extracted has two population sets— the positive and negative—and is defined as:
\[
u_{ij} = (\mathbf{U}_{ij}^{(P)}, \mathbf{U}_{ij}^{(N)})
\]
\[
\begin{array}{cc}
\mathbf{U}_{ij}^{(P)} = (a_{ij}^{(1,P)}, a_{ij}^{(2,P)}, \dots, a_{ij}^{(N_p,P)}) \quad &
\mathbf{U}_{ij}^{(N)} = (a_{ij}^{(1,N)}, a_{ij}^{(2,N)}, \dots, a_{ij}^{(N_n,N)})
\end{array}
\]

\noindent 
where \( \mathbf{U}_{ij}^{(P)} \) represents activation units in the positive set, of size \( N_p \), and \( \mathbf{U}_{ij}^{(N)} \) represents activation values in the negative set, of size \( N_n \). Welch’s t-test is conducted to statistically assess the difference in its responses under the positive and negative contrast conditions:
\[
t_{ij} = \frac{\bar{U}_{ij}^{(P)} - \bar{U}_{ij}^{(N)}}{\sqrt{\frac{S_P^2}{N_p} + \frac{S_N^2}{N_n}}}
\]

\noindent where:
\[
\bar{U}_{ij}^{(P)} = \frac{1}{N_p} \sum_{k=1}^{N_p} a_{ij}^{(k,P)}, \quad \bar{U}_{ij}^{(N)} = \frac{1}{N_n} \sum_{k=1}^{N_n} a_{ij}^{(k,N)}
\]

\[
S_P^2 = \frac{1}{N_p - 1} \sum_{k=1}^{N_p} \left( a_{ij}^{(k,P)} - \bar{U}_{ij}^{(P)} \right)^2, \quad S_N^2 = \frac{1}{N_n - 1} \sum_{k=1}^{N_n} \left( a_{ij}^{(k,N)} - \bar{U}_{ij}^{(N)} \right)^2
\]

\noindent
The t-values \( t_{ij} \), with \( i \in \{1, \dots, M\} \) indexing transformer block and \( j \in \{1, \dots, H\} \) indexing activation units, represent the individual entries of the matrix 
\[
T \in \mathbb{R}^{M \times H},
\]
which quantifies the difference in activation between the positive and negative contrast conditions. This t-value distribution reveals the relative importance of individual activation units under contrasting task conditions, thereby guiding the selection process that is detailed in the next section.

\subsection{Units Selection}

Based on the computed \( t \)-values in the matrix \( T \), activation units are ranked according to their differential responses under the contrasting task conditions. The top \( k\% \) of units—those with the highest \( t \)-values—are considered most strongly associated with the targeted cognitive task, while the bottom \( k\% \) with the lowest \( t \)-values are deemed least involved. To facilitate further analysis, we define a binary mask \( \Gamma \in \{0,1\}^{M \times H} \) that serves as an indicator function for unit selection. Specifically, the mask is defined as:

\[
\Gamma_{ij} =
\begin{cases} 
1, & \text{if unit } u_{ij} \text{ meets the selection criterion}, \\[1ex]
0, & \text{otherwise},
\end{cases}
\]

\noindent
where \( i \in \{1, \dots, M\} \) and \( j \in \{1, \dots, H\} \).  This mask enables us to identify the units most (or least) associated with the targeted cognitive network. Under the random condition, k\% of the activation units are uniformly sampled from the outputs of all transformer blocks (from block 1 to block M).

\section{Localizer Type}
\label{app:loc_type}

\subsection{Theory of Mind Localizer}

The ToM localizer, developed by \citep{dodell-feder_fmri_2011}, distinguishes cognitive engagement by contrasting False-Belief and False-Photograph stories. False-Belief stories, inspired by the classic Sally-Ann test \citep{baron-cohen_does_1985}, describe scenarios in which a character holds incorrect beliefs regarding the actual state of affairs. Conversely, False-Photograph stories do not involve human agents but rather present an outdated representation of a scene. The critical difference between the two lies in the requirement to infer another individual's mental state—a key aspect of ToM. Their method utilizes 10 False-Belief stories contrasted with 10 False-Photograph stories. Examples of both types are illustrated in figure \ref{fig:tom_md_loc}.

\subsection{Multiple Demand Localizer}

\begin{figure}[ht]
    \centering
    \begin{tabular}{cc}
        \multicolumn{2}{c}{\textbf{MD Localizer}} \\[5pt]
        \begin{subfigure}{0.35\textwidth}
          \includegraphics[width=\textwidth]{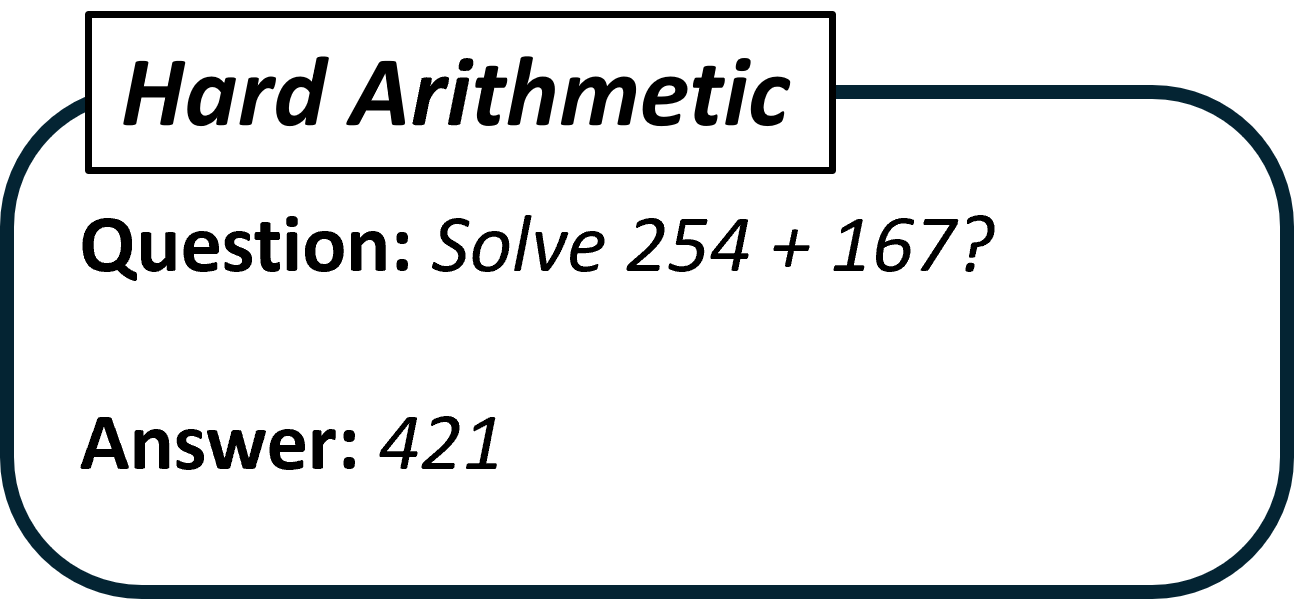}
        \end{subfigure} &
        \begin{subfigure}{0.35\textwidth}
          \includegraphics[width=\textwidth]{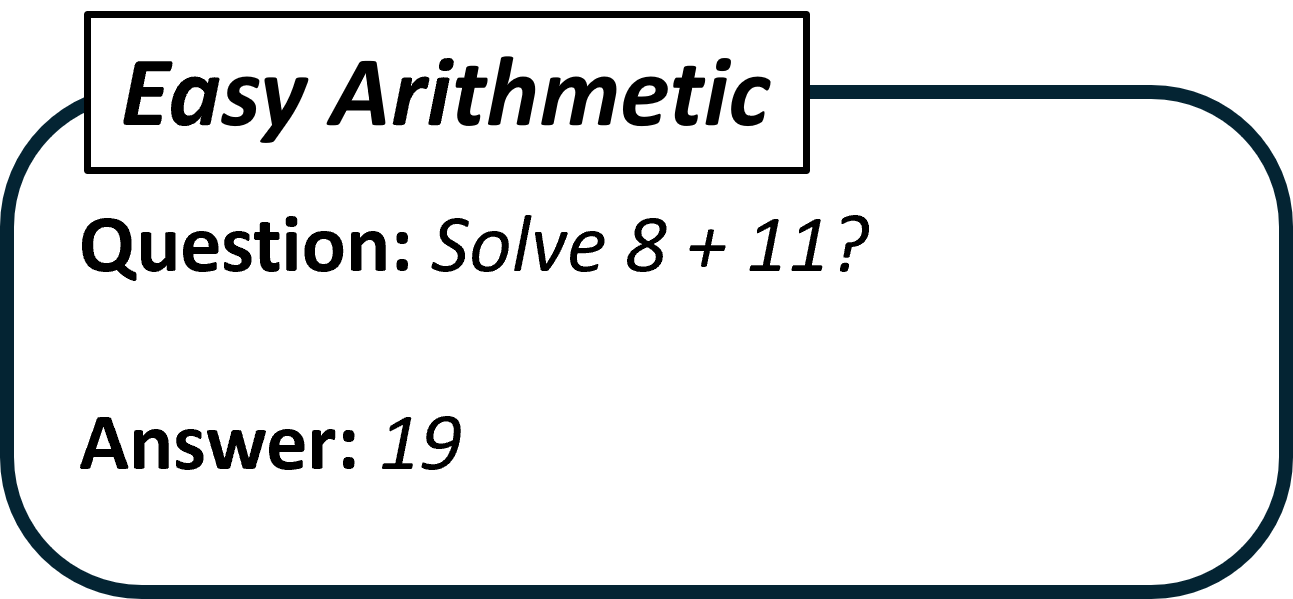}
        \end{subfigure} \\[5pt]
         \multicolumn{2}{c}{\textbf{ToM Localizer}} \\[5pt]
        \begin{subfigure}{0.45\textwidth}
          \includegraphics[width=\textwidth]{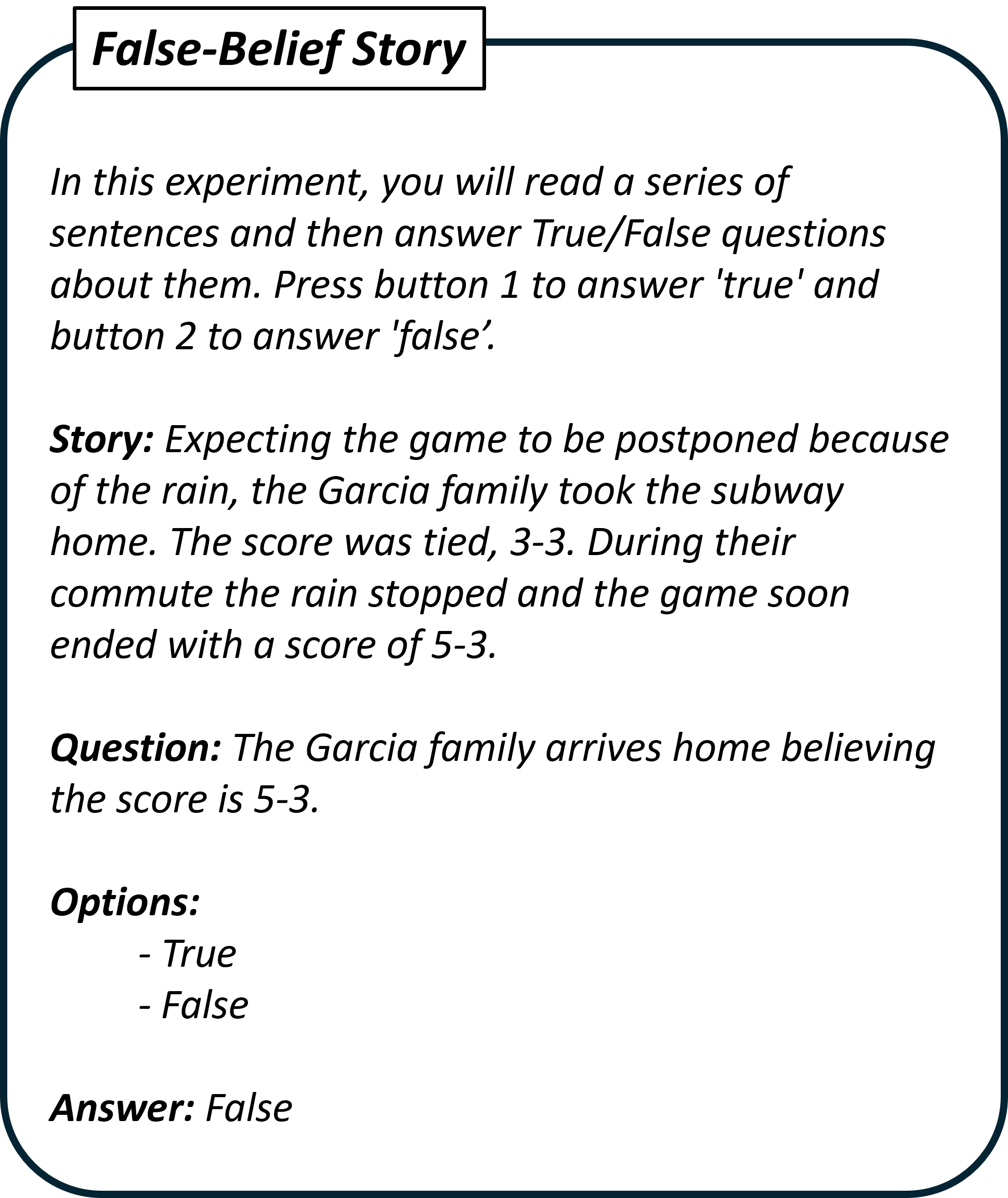}
        \end{subfigure} &
        \begin{subfigure}{0.45\textwidth}
          \includegraphics[width=\textwidth]{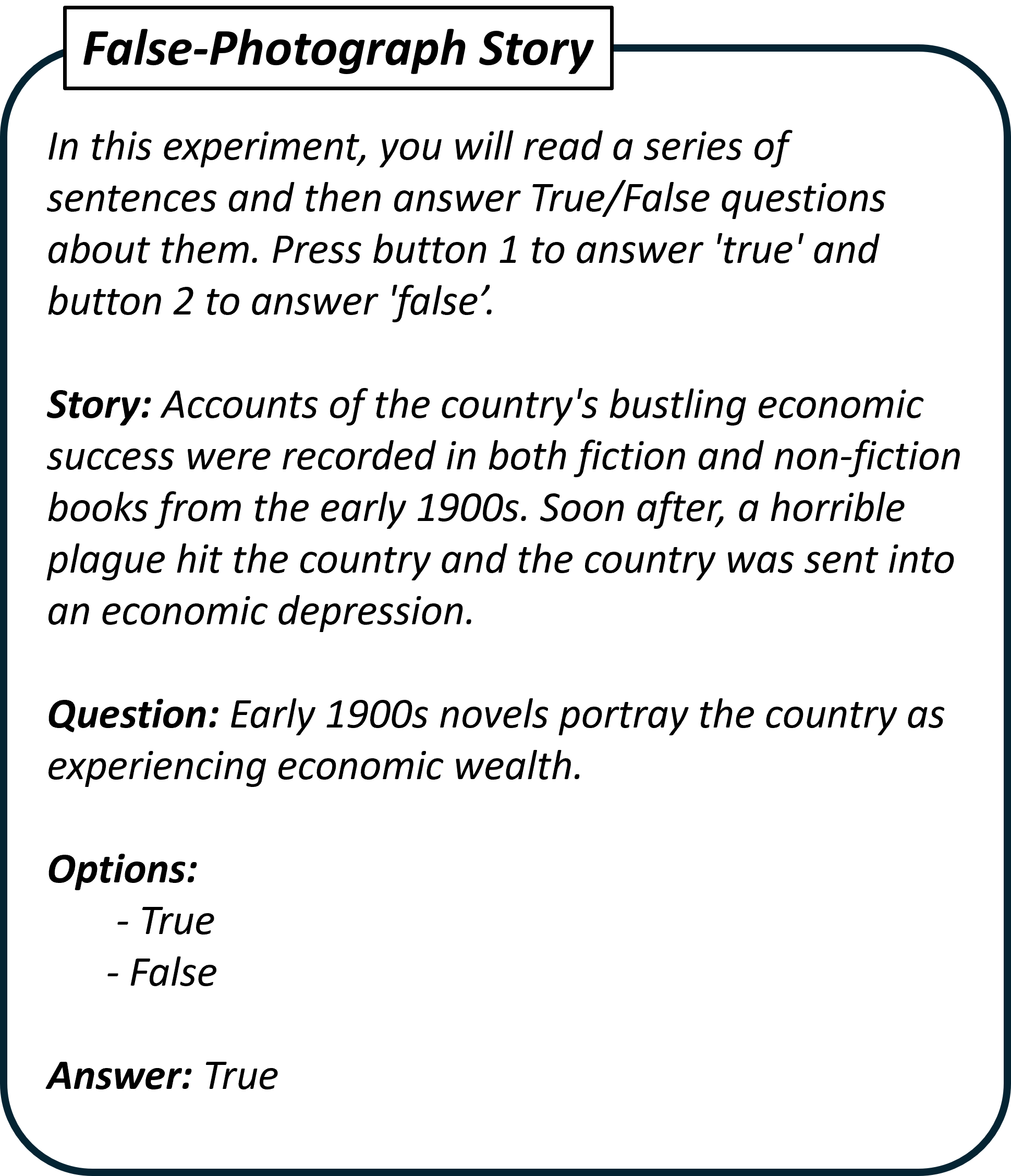}
        \end{subfigure} \\
    \end{tabular}
    \caption{\textbf{Example of positive and negative localizer stimuli.} For the MD localizer, the Hard Arithmetic condition involves more challenging addition or subtraction problems with larger numbers, while the Easy Arithmetic condition employs simpler arithmetic with smaller numbers; each condition comprises 100 stimuli. For the ToM localizer, two types of short stories are presented—False-Belief (left) and False-Photograph (right)—with 10 stimuli per condition.}
    \label{fig:tom_md_loc}
\end{figure}

For this study, we adopt the arithmetic MD localizer introduced by \citep{alkhamissi_llm_2024}, which presents arithmetic problems in a verbal format. In the easy condition, participants solve addition and subtraction problems involving small numbers, whereas the hard condition requires the same operations with larger numbers. In the hard condition (positive set), arithmetic problems were created by randomly selecting two integers between 100 and 200, with the operation—either addition or subtraction—also determined at random. Conversely, in the easy condition (negative set), two integers between 1 and 20 were sampled, and the arithmetic operation was similarly chosen at random. Each condition comprised a set of 100 stimuli. An illustrative example of this arithmetic contrast is shown in figure \ref{fig:tom_md_loc}.

\section{Benchmarks}
\label{app:benchmark}

To ensure consistency across different benchmarks, candidate answers are enumerated with letter labels, and the model is prompted to select the correct answer by generating the corresponding letter rather than the full response. This approach is necessary because answer formats vary across benchmarks. For example, in datasets such as ToMi and OpenToM, candidate answers are typically short—consisting of one or two tokens—whereas FanToM uses full sentences as answer choices, and MD datasets often include candidate options that consist of mathematical formulas or code snippets written in a formal notation rather than natural language. By enforcing a one-token response format, we maintain homogeneity across datasets and ensure that all answers can be processed consistently. Additionally, this method optimizes inference efficiency; while prompt lengths may vary depending on the dataset, restricting the model’s prediction to a single token significantly accelerates response generation. The following sections provide further details on these ToM and MD datasets.

\subsection{Theory of Mind Benchmarks}

Given the inherent complexity of ToM reasoning tasks, our evaluation focuses exclusively on first-order false-belief questions. In first-order tasks, an agent holds a belief about the world that may be either true or false (for example, Sally may believe her toy remains in the basket despite it having been moved). By contrast, second-order tasks require an agent to infer another individual’s belief (e.g., Sally thinks that Anne believes the toy is in the basket). Recent studies on LLMs reveal that these models handle first-order false-belief questions significantly better than more complex second-order questions \citep{kim_fantom_2023, van_duijn_theory_2023, sclar_minding_2023}. Given that first-order tasks already present a substantial challenge under our experimental setup, they serve as a robust baseline for assessing ToM-like reasoning in LLMs.

\textbf{ToMi} is generated using a stochastic, rule-based algorithm, drawing inspiration from the Sally-Ann Test. The story structure involves two characters, an object that is relocated, and two locations—one where the object originates and another where it is moved. These elements are organized into a narrative, followed by a question asking each character where they believe the object is located \citep{le_revisiting_2019}, corresponding to both false-belief and true-belief scenarios. The ToMi dataset used in this study was preprocessed by \citet{sap_neural_2022}. In our analysis, we consider the 231 false-belief questions.

\noindent
\textbf{OpenToM} is designed to assess false-belief capabilities in LLMs \citep{xu_opentom_2024}. Each story involves two characters, an entity of interest, and multiple locations or containers where the entity may be placed. The dataset is constructed using a human-in-the-loop process, where GPT-3.5-Turbo generates initial narratives, which are subsequently refined through human annotation. This methodology ensures the incorporation of character personality traits, intentions, and actions, creating more realistic and contextually grounded interactions. The datasets comprises 686 false-belief questions.

\noindent
\textbf{FanToM} evaluates ToM reasoning in multi-party conversations with information asymmetry, consisting of 256 conversations across various topics where characters enter and leave, creating knowledge gaps \citep{kim_fantom_2023}. The benchmark assesses LLMs' ability to track characters' beliefs through short and full conversations. The dataset comprises 642 questions.

\subsection{Multiple Demand Benchmark}
    \label{sub:md_bench}

The assessment of MD is conducted using a mathematical reasoning task, as this network is associated with domain-general cognitive control and problem-solving processes \citep{assem_domain-general_2020, woolgar_fluid_2010}. 

\textbf{MATH} is a unimodal mathematical reasoning benchmark comprising 4,914 questions covering various topics, including Geometry, Counting \& Probability, Calculus, and Algebra \citep{hendrycks_measuring_2021}. Each question is categorized into five levels of difficulty. In this work, we use the four-choice multiple-answer format introduced by \citep{zhang_multiple-choice_2024}.

\textbf{MMStar} is a multimodal benchmark that rigorously evaluates VLMs on their ability to integrate and reason with both visual and textual information. Comprising 1,500 questions, the dataset is designed so that correct answers are derived from an analysis of the visual content, covering various topics that assess advanced multimodal reasoning. These topics include coarse and fine-grained perception, instance reasoning, logical deduction, and domain-specific knowledge in science, technology, and mathematics.

\textbf{MathVista} is a multimodal benchmark designed to evaluate the mathematical reasoning abilities—from algebra and geometry to logical and statistical reasoning—of VLMs within visual contexts. In our study, we exclusively examine the miniTest subset of MathVista—a segment that replicates the distribution of the full dataset while offering computational efficiency. To streamline our evaluation, we further restrict our analysis to questions that include candidate answer options, resulting in a refined set of 540 questions.

\end{document}